\documentclass[10pt,twocolumn,letterpaper]{article}

\usepackage[pagenumbers]{iccv} 

\usepackage{graphicx}	
\usepackage{amsmath}	
\usepackage{amssymb}	
\usepackage{booktabs}
\usepackage{times}
\usepackage{microtype}
\usepackage{epsfig}
\usepackage{caption}
\usepackage{float}
\usepackage{placeins}
\usepackage{color, colortbl}
\usepackage{stfloats}
\usepackage{enumitem}
\usepackage{tabularx}
\usepackage{xstring}
\usepackage{multirow}
\usepackage{xspace}
\usepackage{url}
\usepackage{subcaption}
\usepackage{xcolor}
\usepackage[hang,flushmargin]{footmisc}
\usepackage{standalone}

\definecolor{iccvblue}{rgb}{0.21,0.49,0.74}
\usepackage[pagebackref,breaklinks,colorlinks,allcolors=iccvblue]{hyperref}

\title{InstructRestore: Region-Customized Image Restoration with Human Instructions}

\author{Shuaizheng Liu$^{1,2}$, Jianqi Ma$^{1}$, Lingchen Sun$^{1,2}$, Xiangtao Kong$^{1,2}$,  Lei Zhang$^{1,2,\dagger}$ \\
{$^{1}$The Hong Kong Polytechnic University \qquad $^{2}$OPPO Research Institute} \\
{\tt\small shuaizhengliu21@gmail.com, jianqi.ma@connect.polyu.hk \ ling-chen.sun@connect.polyu.hk} \\
{\tt\small xiangtao.kong@connect.polyu.hk \ cslzhang@comp.polyu.edu.hk}
\\
}

\begin{document}
\maketitle
\begin{abstract}

Despite the significant progress in diffusion prior-based image restoration, most existing methods apply uniform processing to the entire image, lacking the capability to perform region-customized image restoration according to user instructions. 
In this work, we propose a new framework, namely \textbf{InstructRestore}, to perform region-adjustable image restoration following human instructions.
To achieve this, we first develop a data generation engine to produce training triplets, each consisting of a high-quality image, the target region description, and the corresponding region mask. With this engine and careful data screening, we construct a comprehensive dataset comprising 536,945 triplets to support the training and evaluation of this task.
We then examine how to integrate the low-quality image features under the ControlNet architecture to adjust the degree of image details enhancement. 
Consequently, we develop a ControlNet-like model to identify the target region and allocate different integration scales to the target and surrounding regions, enabling region-customized image restoration that aligns with user instructions.
Experimental results demonstrate that our proposed InstructRestore approach enables effective human-instructed image restoration, such as images with bokeh effects and user-instructed local enhancement. Our work advances the investigation of interactive image restoration and enhancement techniques. 
Data, code, and models will be found at \url{https://github.com/shuaizhengliu/InstructRestore.git}
\end{abstract}

\renewcommand{\thefootnote}{}
\footnotetext{$\dagger$ Corresponding author. This work is supported by the PolyU-OPPO Joint Innovative Research Center.}

\section{Introduction}
\label{sec:intro}

\begin{figure*}[t]
  \centering
  \includegraphics[width=\linewidth]{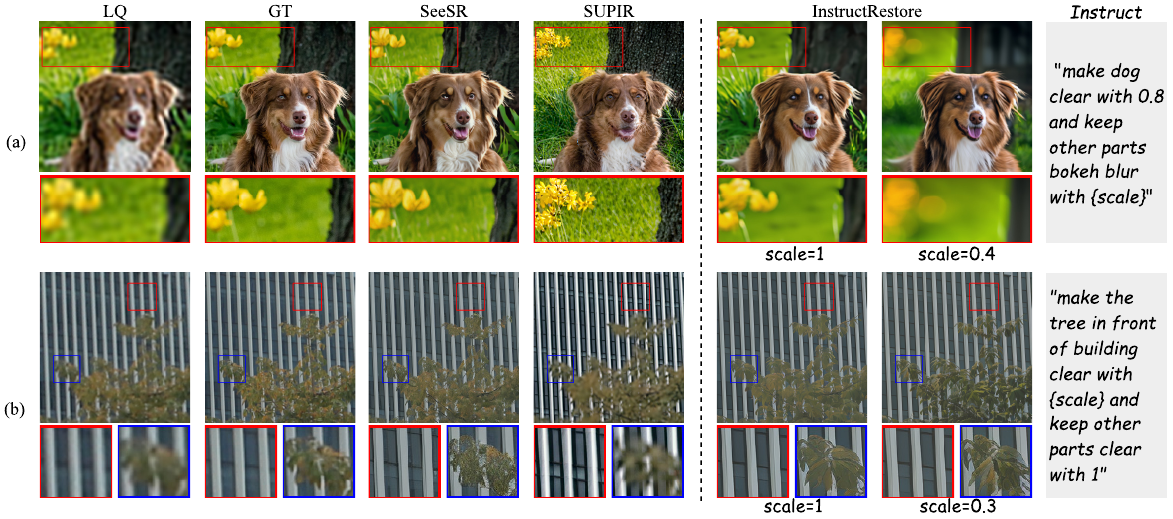}
  \vspace{-7mm}
  \caption{Our proposed \textbf{InstructionRestore} framework enables region-customized restoration following human instruction. As shown in (a), current methods \cite{wu2024seesr,yu2024scaling} tend to incorrectly restore the bokeh blur, while  method allows for adjustable control over the degree of blur based on user instructions. In (b), existing methods fail to achieve region-specific enhancement intensities, while our approach can simultaneously suppress the over-enhancement in areas of building and improve the visual quality in areas of leaves.}
  \label{fig:Intro_fig}
  \vspace{-4mm}
\end{figure*}

Image restoration (IR) is a fundamental problem in computer vision, with the aim of recovering high-quality images from degraded inputs. In the past decades, deep learning-based discriminative IR models~\cite{liang2021swinir, lu2022transformer, chen2022simple, dong2015image, sun2024perception} have made significant progress with paired training data. However, most of them employ fidelity-oriented loss functions to optimize the network, resulting in over-smoothed outputs with insufficient details. To address this issue, researchers have developed models based on Generative Adversarial Networks (GANs)~\cite{park2023content, wang2018esrgan, wang2021real, liang2022details}, which align the distribution of IR outputs with that of high-quality images. Although these approaches can generate realistic details and improve perception quality, they will also introduce many visual artifacts.
Recently, with the advent of pretrained text-to-image (T2I) generation models such as Stable Diffusion (SD) \cite{rombach2022high}, which can more effectively model the complex distribution of natural images, researchers have started to explore the use of powerful SD priors to produce realistic IR outcomes \cite{wang2024exploiting,lin2024diffbir, yang2024pixel, wu2024seesr, yu2024scaling, wu2025one, sun2024pisasr, wang2024sinsr}. 

By generating details semantically consistent with the underlying content of the image, SD-based methods can improve significantly over GAN-based methods in synthesizing realistic image details.
However, existing methods perform restoration uniformly on the whole input image, lacking the ability to adjust region-customized restoration levels according to image content and user instructions.
For example, in the photography of targeted objects (\eg, portrait), the background is often blurred to enhance the aesthetic appeal of the foreground region. When applying to such images, existing generative prior-based IR methods may produce many unnecessary textures/details on the background regions, as shown in Fig.~\ref{fig:Intro_fig}(a). 
In addition, different regions of an image require different degrees of restoration to achieve the best overall perceptual quality. 
For example, textured regions often require a higher level of generative details to achieve sharper visual appearance, while plane regions need a lower degree of generation for visual smoothness, as illustrated in Fig.~\ref{fig:Intro_fig}(b).
Unfortunately, existing methods make it difficult to achieve such customized restoration of different regions. 

To address the limitation mentioned above, we propose a new framework, namely \textbf{InstructRestore}, enabling users to specify the target region, as well as the desired degree of restoration, through natural language instructions. Our InstructRestore approach can follow the user instructions to precisely restore the target regions while properly handling the restoration of the background areas.
To begin with this novel task, 
we need a dataset for training and evaluation, which should offer descriptions of target regions to construct human instruction, along with corresponding region masks.
To the best of our knowledge, there is not a publicly available dataset that provides such triplets of high-quality images, referring descriptions, and the corresponding region masks.
The most relevant datasets to our task can be the referential segmentation datasets such as RefCOCO~\cite{yu2016modeling}. However, its image quality and resolution are insufficient to support IR tasks. 
To bridge this gap, we develop a data generation engine. Utilizing Semantic-Sam~\cite{li2024segment} and Osprey~\cite{yuan2024osprey} models, we obtain masks and initial descriptions from a set of selected high-quality images.
We then use large language models (LLMs), more specifically Qwen~\cite{yang2024qwen2}, to iteratively parse and refine these descriptions, formatting them to meet the instructional requirements of IR tasks. Finally, we build a dataset of 536,945 triplets, covering diverse scenes such as plants, buildings, animals, \etc.

Building upon this dataset, we train the InstructRestore model for region-customized IR with user instructions. To ensure that the model can accurately identify the human-specified region and properly enhance the designated area, we propose integrating the conditional features of low-quality input images into a ControlNet-like architecture. 
Instructions are used as text prompts to the control-branch of ControlNet. Trained on our curated dataset, the control-branch could simultaneously generate region masks and conditional features. By applying distinct integration scales to the conditional features of user-customized regions and their surroundings, our InstructRestore model achieves locally controlled restoration that aligns with user intentions.

Our key contributions are summarized as follows. 
(1) First, we introduce the task of region-customized image restoration with human instruction, which represents an important class of practical IR tasks.
(2) Second, we develop a data generation engine and construct a large-scale dataset with $536,945$ triplets to support this task.
(3) Finally, we design the InstructRestore model, which understands user instructions for region-customized restoration. Our experiments demonstrate the capability and effectiveness of our InstructRestore model, showcasing its great potential for interactive and user-instructed image restoration.

\vspace{-5pt}
\section{Related Work}
\label{sec:related}
\textbf{Diffusion-based Restoration in the Wild}.
Recent diffusion models have significantly advanced the task of IR in the wild, addressing mixed degradations such as noise, blur, JPEG compression, and resolution reduction. StableSR~\cite{wang2024exploiting} and DiffBIR~\cite{lin2024diffbir} treat the low-quality (LQ) input as condition to guide reverse diffusion process. PASD~\cite{yang2024pixel} and SeeSR~\cite{wu2024seesr} introduce the semantic prompts like short captions or tags to enrich the result with finer semantic details. SUPIR~\cite{yu2024scaling} scales up datasets along with long descriptions to boost perceptual quality with SDXL~\cite{podell2023sdxl} pre-trained model. To tackle the inefficiency of iterative sampling, one-step diffusion methods~\cite{wu2025one, sun2024pisasr} have emerged, ensuring quality with faster inference. Despite their advancements, existing methods perform restoration uniformly, failing to accommodate user preferences for region-specific refinements.

\noindent \textbf{Instruction-guided Editing and Restoration}.
Natural language instructions enable intuitive human-AI collaboration by translating high-level intent into pixel-level operations. Instruction-guided image editing methods like InstructPix2Pix~\cite{brooks2023instructpix2pix} and MagicBrush~\cite{zhang2024magicbrush} have demonstrated remarkable capabilities in spatially aware manipulations. Subsequent works like MGIE~\cite{fu2023guiding} and SmartEdit~\cite{huang2024smartedit} further advance instruction comprehension through multimodal LLMs. Others~\cite{li2024zone, guo2024focus} focus on region-specific control, ensuring editing explicitly defined areas by user instructions. However, these breakthroughs remain confined to semantic-level manipulation rather than physically grounded restoration. To address this, recent efforts have incorporated user instructions into restoration frameworks. InstructIR~\cite{conde2024instructir} and PromptFix~\cite{yu2024promptfix} leverage task-specific instructions to enable a single model to handle multiple restoration tasks, including denoising, deblur, rain removal, \etc. SPIRE~\cite{qi2024spire} incorporates semantic descriptions to handle in-the-wild restoration scenarios. However, these methods primarily use instructions for task differentiation or global parameter tuning, lacking the ability to perform region-specific refinements. Our work introduces the first instruction-guided restoration framework that enables region-specific refinements through natural language commands, addressing the critical limitation of global-only operations in prior arts.

\section{Dataset Construction}
\label{sec:dataset}

\begin{figure*}[t]
  \centering
  \includegraphics[width=1.0\linewidth]{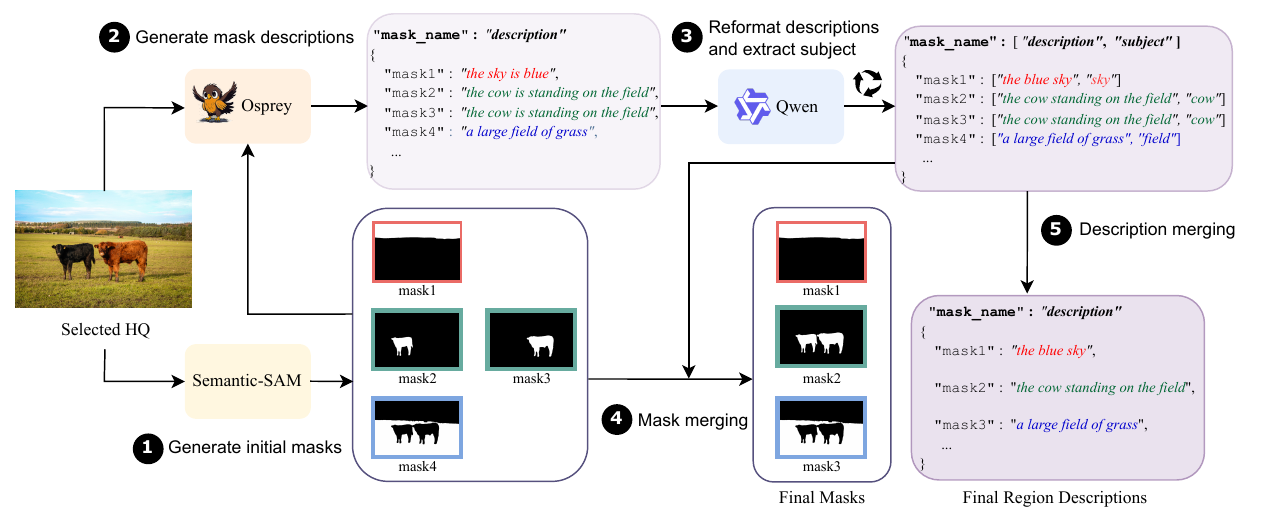}
\vspace{-9mm}
  \caption{Illustration of the annotation pipeline. For selected high-quality images, Semantic-SAM~\cite{li2024segment} generates initial masks, followed by Osprey~\cite{yuan2024osprey} for region-level descriptions. Qwen~\cite{yang2024qwen2} reformats descriptions into noun phrases and extracts semantic subjects. Identical semantics are merged to produce final masks and region captions.}
  \label{fig:datacollection}
  \vspace{-4mm}
\end{figure*}

InstructRestore aims to restore target regions following user-provided instructions while appropriately restoring background areas. To achieve this goal, the model needs to understand the semantic information of the target regions for performing localized restoration. 
A critical requirement for training such a model is the availability of a large-scale dataset, which simultaneously offers high-quality images, detailed descriptions of target regions, and corresponding region masks.
In this paper, we develop a data generation engine to build such a comprehensive dataset to facilitate the research of InstructRestore tasks. The data generation process is detailed in Fig.~\ref{fig:datacollection}.

\subsection{Dataset Construction Pipeline}
\noindent \textbf{High-quality ground-truth image collection.}
High-resolution and high-quality ground-truth (GT) images are critical for training IR models. Therefore, we collect high-quality images from LSDIR~\cite{li2023lsdir}, EntitySeg~\cite{Qi_2023_EntitySeg} train set and EBB!~\cite{ignatov2020rendering} bokeh train set with shorter side larger than $512$ pixels and MUSIQ~\cite{ke2021musiq} score larger than $60$.

\begin{figure}[t]
 \centering
  \includegraphics[width=0.8\linewidth]{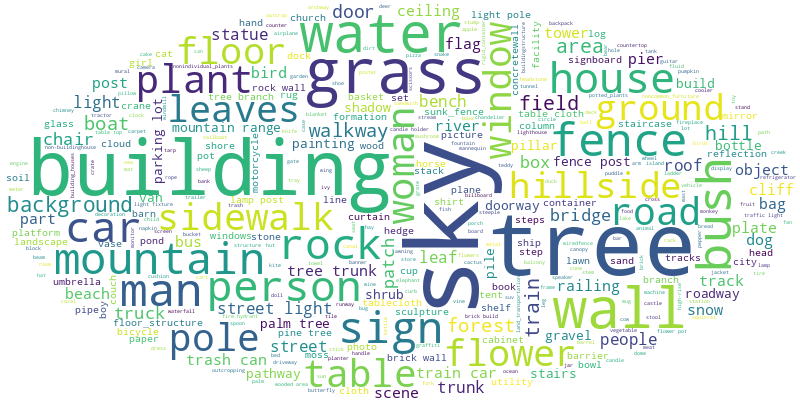}
  \vspace{-3mm}
  \caption{Subject distribution word cloud generated from region captions in our dataset. Word size corresponds to the relative frequency of extracted subjects.}
  \label{fig:world cloud}
  \vspace{-4mm}
\end{figure}

\noindent \textbf{Annotation pipeline.}
To obtain high-quality images, we design an automatic annotation pipeline to extract the semantic region masks and their corresponding descriptions with a combination of state-of-the-art models. 
In the mask extraction phase, we first utilize a state-of-the-art segmentation model, \eg, Semantic-SAM~\cite{li2024segment}, to generate coarse-grained semantic segmentation masks for the semantic region of the images. For images from EntitySeg~\cite{Qi_2023_EntitySeg}, we directly reuse their pre-annotated masks. Once the masks are obtained, we pair each image with its mask and feed them into a multi-modal large language model,  Osprey~\cite{yuan2024osprey}, to generate region-level descriptions. These descriptions serve as part of instructions to specify the regions to be processed or restored. At this stage, though we obtain preliminary masks and descriptions, they are still far from perfect as our training data due to two key issues below: (1) Semantic-SAM~\cite{li2024segment} occasionally produces multiple mask pieces for one semantic meaning, leading region ambiguity and harmful for the region customization learning; (2) the descriptions are not always in noun phrase format due to the response arbitrariness, making them unsuitable for embedding into instructions.

To address these issues, we first utilize Qwen-7B~\cite{qwen}, a large language model (LLM), to perform the following two tasks through prompt tuning: (1) parsing the subject from the descriptions and (2) reformatting them into noun phrases. Due to the instability in the LLM's output, \eg, repetition or spelling errors, we iteratively perform the refinement process. Specifically, we identify error cases and re-execute the above process by a larger LLM, Qwen-72B~\cite{qwen}. This cycle is repeated 3 times to ensure high-quality outputs. More details can be found in the \textbf{supplementary file}. Finally, based on the parsed subject, we merge the masks and their corresponding descriptions for regions with identical semantics.

\subsection{Dataset Statistics}
\begin{table}
\centering
\caption{Statistics of our dataset and related datasets.}
\vspace{-3mm}
\label{tab:comparison of dataset}
\resizebox{\columnwidth}{!}{ 
\begin{tabular}{c c c c c}
\hline
Datasets & \shortstack{Annotation \\ Amount}  & \shortstack{Min \\ Resolution} & \shortstack{Max \\ Resolution} &  MUSIQ  \\ \hline
RefClef~\cite{kazemzadeh2014referitgame} & 99,523 & 320$\times$480 & 360$\times$480 & 67.06  \\
RefCOCO~\cite{yu2016modeling} & 196,771 & 157$\times$160 & 640$\times$637 & 69.73 \\
RefCOCO+~\cite{yu2016modeling} & 196,737 &157$\times$160 & 640$\times$637 & 69.73 \\
RefCOCOg~\cite{mao2016generation} & 208,960 & 157$\times$160 & 640$\times$637 & 69.73 \\ \hline
Ours & 536,945  & 540$\times$540 & 4464$\times$2244 & 71.87 \\

\hline
\end{tabular}
}
\vspace{-4mm}
\end{table}
As illustrated in Fig.~\ref{fig:datacollection}, our dataset provides triplets of high-quality GT images, region masks, and descriptive captions. To underscore the relevance and utility of our dataset, we compare it with the most relevant referential segmentation datasets including RefClef~\cite{kazemzadeh2014referitgame}, RefCOCO~\cite{yu2016modeling}, RefCOCO+~\cite{yu2016modeling} and RefCOCOg~\cite{mao2016generation} in Table~\ref{tab:comparison of dataset}. These datasets also provide masks and captions for semantic regions. The comparison focuses on the number of annotations, the range of image resolutions, and MUSIQ-based quality scores.

As can be seen from Table~\ref{tab:comparison of dataset}, existing datasets, while widely used for segmentation, exhibit critical limitations for IR tasks. Their images are capped at resolution less than 650 pixels and their MUSIQ scores fall significantly below ours. In contrast, our dataset not only provides 536,945 annotated regions (surpassing other datasets in scale) but also delivers higher-resolution images with superior perceptual quality, meeting the need for IR tasks. Our dataset enables both precise semantic control and photorealistic restoration. To further illustrate the semantic diversity and applicability of our dataset, we plot a word cloud based on the frequency of subjects extracted from the annotation descriptions. The size of each word in the cloud corresponds to its relative frequency, and larger words represent more prevalent subjects. As shown in Fig.~\ref{fig:world cloud}, our dataset encompasses a wide range of common semantic categories, including animals, plants, natural landscapes, pedestrians, and other subjects frequently encountered in restoration.

\section{InstructRestore Model Design}
\label{sec:method}


\begin{figure*}[t]
  \centering
  \includegraphics[width=\linewidth]{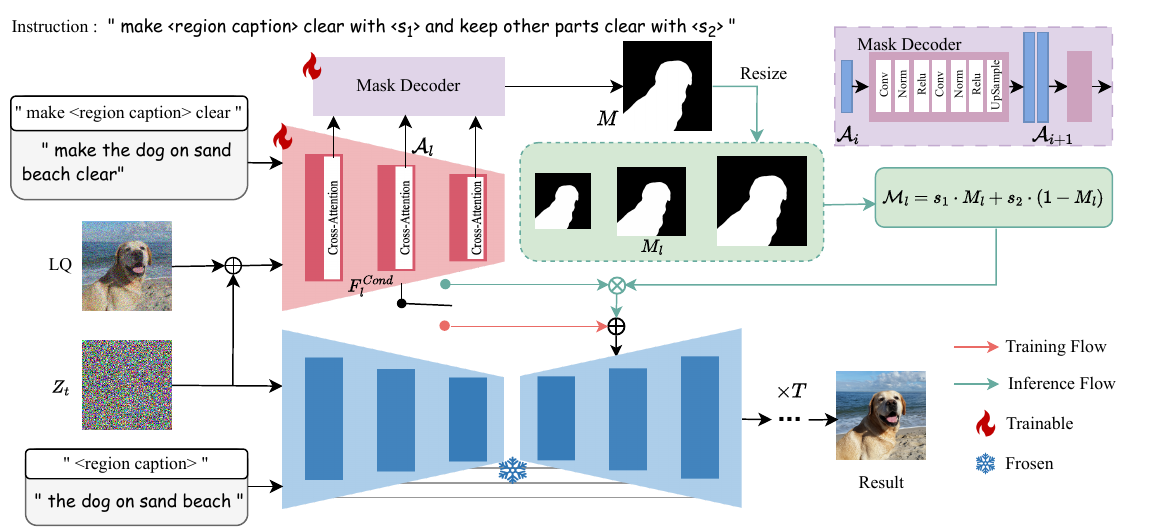}
  \vspace{-4mm}
  \caption{Framework of InsturctRestore. The framework uses red and green arrows to denote training and inference processes respectively. During testing, user instructions are parsed to generate target-region semantic masks, with differentiated coefficient modulation applied to conditional features inside/outside mask regions, enabling instruction-guided region-adaptive restoration effects.}
  \label{fig:Framework}
  \vspace{-4mm}
\end{figure*}

\subsection{Training Framework}
Existing SD-based IR models often employ a ControlNet architecture with the low-quality (LQ) image as a conditional signal. 
We observe that scaling the ControlNet features by a coefficient $\alpha$ during inference provides flexible control of the data fidelity and semantic enhancement. 
Building on this insight, we propose InstructRestore, a framework that first identifies the regions specified in user instructions and then performs region-customized restoration.

\noindent \textbf{Architecture design}.
As shown in Fig. \ref{fig:Framework}, our InstructRestore model consists of a pre-trained SD backbone, the ControlNet, and a lightweight mask decoder. The pre-trained SD model is frozen during the entire training stage.
The region captions $c_{R}$ extracted from the user instructions $c_I$ act as text prompts for the SD model, providing semantic guidance and to generate semantic details.
ControlNet duplicates the encoder and middle blocks of the pre-trained SD UNet as trainable copies. It receives features extracted from the LQ image and user instructions $c_I$ as input, then extracts hierarchical conditional features from the input and injects them into the SD UNet decoder blocks at multiple scales.

To accurately localize the target regions in user instructions, we design a mask decoder to predict a spatial mask $\hat{M}$. Since ControlNet is initialized from the pre-trained SD UNet, it has been revealed ~\cite{hertz2022prompt} that the cross-attention features between textual and visual embeddings exhibit strong responses to text-described semantic regions. We then extract cross-attention features $\{\mathcal{A}_l\}_{l=1}^L$ between textual embeddings and visual features at each scale of the ControlNet as input to the Mask decoder, which is designed with a pyramidal structure to effectively process multi-scale features. The features of each scale $\mathcal{A}_l$ are first passed through two blocks, each consisting of a convolutional layer~(Conv), group normalization~(GN), and a ReLU activation. The processed features are then upsampled and concatenated with the cross-attention features $\mathcal{A}_{l+1}$ at the next scale. The combined features are processed by another Conv-GN-ReLU block and passed to the subsequent scale.

\noindent\textbf{Training process}.
Our constructed dataset consist of triplets $[I_{\text{HQ}}, M, c_R]$, where $I_{\text{HQ}}$ denotes the high-quality (HQ) GT image, $M$ is the binary mask specifying the target region, and $c_R$ is the textual region caption of the masked area. During model training, we first apply the Real-ESRGAN degradation model $\mathcal{D}$ to $I_{\text{HQ}}$ to generate the LQ input $I_{\text{LQ}} = \mathcal{D}(I_{\text{HQ}})$. The region caption $c_R$ is utilized to construct task-specific training instructions $c_I$ based on the restoration objective. For region-customized restoration, the instruction template is \textit{``make $\{c_R\}$ clear"}, while $\{c_R\}$ is extracted from the region captions in the training data. For bokeh-aware restoration, the template becomes \textit{``make $\{c_R\}$ clear and keep other parts bokeh blur"}. These instructions $c_I$ are sent into the ControlNet branch, while the text prompt fed into the SD backbone is \textit{``$\{c_R\}$ in front of bokeh background"}.

The HQ image $I_{\text{HQ}}$ is first encoded into the latent space by the pre-trained VAE encoder, yielding $z_0$. The diffusion process progressively corrupts $z_0$ with Gaussian noise over randomly sampled timesteps $t$, resulting in noisy latent states $z_t = \sqrt{\alpha_t} z_0 + \sqrt{1 - \alpha_t} \epsilon$, where $\epsilon \sim \mathcal{N}(0, \mathbf{I})$ and $\alpha_t$ follow a cosine noise schedule. The ControlNet takes $I_{\text{LQ}}$ and $c_I$ as input to produce conditional features $\{f_l^{\text{cond}}\}_{l=1}^L$, which are added to the frozen SD UNet decoder with scaling coefficient $\alpha =1$ during training. The training flow is highlighted in red in Fig. \ref{fig:Framework}.
The mask decoder takes cross-attention features $\{\mathcal{A}_l\}_{l=1}^L$ from ControlNet as input to generate target region masks $\hat{M}$, supervised by the GT masks $M$ with Cross-Entropy loss. The InstructRestore network, denoted by $\epsilon_\theta$, is conditioned on noisy latent $z_t$, LQ images $I_{\text{LQ}}$, instruction $c_I$, and region caption $c_R$. The training objective $\mathcal{L}$ combines the standard diffusion loss and mask supervision $\mathcal{L}_{\text{mask}}$:
\begin{equation}
\mathcal{L} = \mathbb{E}_{t, \epsilon} \left[ \| \epsilon - \epsilon_\theta(z_t, t, I_{\text{LQ}}, c_I, c_R) \|_2^2 \right] + \lambda \mathcal{L}_{\text{mask}}(\hat{M}, M),
\end{equation}
where $\mathcal{L}_{\text{mask}}(\hat{M}, M) = \text{CrossEntropy}(\hat{M}, M)$ and parameter
$\lambda$ balances the two terms.

\subsection{Region-customized Inference}
After training, our framework enables users to specify target regions and restoration intensities through structured instructions during inference, as shown in the green flow in Fig. \ref{fig:Framework}. The user instructions follow task-specific templates. For general restoration, we set the template as \textit{``make \{region caption\} clear with \{s\textsubscript{1}\}, and make other parts with \{s\textsubscript{2}\}"}; for bokeh-aware restoration, the template is designed as \textit{``make \{region caption\} clear with \{s\textsubscript{1}\}, and keep other parts bokeh blur with \{s\textsubscript{2}\}"}. The \{region caption\} specifies the textual caption of region of interest (\eg, ``the dog on the sand beach"), and $s_1, s_2 \in \mathbb{R}^+$ define the enhancement scales for the target and background regions, respectively. The instruction is parsed as follows: 1) For region-customized restoration, the \textit{``\{region caption\}"} is used as the condition of SD, while for bokeh-aware restoration, \textit{``\{region caption\} in front of bokeh background"} is used; 
2) The main body of instruction is adapted to the task: for region-customized restoration, \textit{``make \{region caption\} clear"} is fed into ControlNet’s text encoder, whereas for bokeh-aware restoration, \textit{``make \{region caption\} clear and keep other parts bokeh blur"} is used; and 3) the fidelity scales $s_1, s_2$ are extracted via regular expression parsing for mask modulation.

The trained ControlNet branch processes the degraded image and the parsed instruction to generate a restoration mask $M \in [0,1]$, which is dynamically resized to match the spatial dimensions of each U-Net upsampling decoder layer, producing masks at multiple scales $\{M_l\}_{l=1}^L$. At each layer $l$, a modulation map is computed as:
\begin{equation}
\mathcal{M}_l = s_1 \cdot M_l + s_2 \cdot (1 - M_l), 
\end{equation}
where $s_1$ and $s_2$ control the enhancement intensities for the target and background regions, respectively. The modulated ControlNet features $F_l^{\text{cond}}$ are fused with the base SD features $F_l^{\text{sd}}$ via element-wise multiplication:
\begin{equation}
F_l^{\text{out}} = F_l^{\text{sd}} + \mathcal{M}_l \odot F_l^{\text{cond}}.
\end{equation}
By applying this modulation progressively across all decoder layers, our framework ensures precise alignment with user intent, \ie, enhancing target regions with intensity $s_1$ while maintaining natural fidelity in non-target areas with intensity $s_2$. Our architecture guarantees seamless transitions between differently enhanced regions, producing photorealistic restoration results that faithfully follow user instructions.

\section{Experiments}
\label{sec:experiment}

\subsection{Experiment Settings}
\textbf{Training details.}
Our method is built on SD2.1~\cite{rombach2022high}. Training data is generated by the data generation engine described in Section~\ref{sec:dataset}. The LQ images are obtained by the Real-ESRGAN~\cite{wang2021real} degradation pipeline. The LQ images and instructions serve as inputs to the model, while the GT images and region masks provide supervision. 
Our model is first trained on the general degradation dataset for 120K iterations, guided by the instruction template \textit{``Make the $\{$ region caption $\}$ clear "}. The training continues by combining the bokeh dataset with the general degradation dataset for 14k iterations. During this stage, the sampling probability is set to $25\%$ for the general degradation dataset and $75\%$ for the bokeh dataset, which is paired with the instruction template \textit{``Make the $\{$ region caption $\}$ clear and keep other parts bokeh blur."}. The training is conducted on two A100 GPUs with a batch size of 64 and an initial learning rate of $5e^{-5}$. AdamW is adopted as the optimizer for network training. 

\noindent\textbf{Comparison methods.}  
As the first instruction-based region-customized IR approach, InstructRestore mainly benchmarks against: (1) GAN-based Real-ESRGAN~\cite{wang2021real}; (2) Diffusion-based methods like StableSR~\cite{wang2024exploiting}, DiffBIR~\cite{lin2024diffbir}, PASD~\cite{yang2024pixel}, SeeSR~\cite{wu2024seesr}, SUPIR~\cite{yu2024scaling}, and OSEDiff~\cite{wu2024one}.

\subsection{Results on Localized Enhancement}
We first show InstructRestore's results with user instructions. Then we demonstrate its precise restoration of specified regions. Finally, we compare it with existing methods.

\noindent \textbf{Test dataset}.
We curate 100 real-world images from multiple sources, including RealSR~\cite{cai2019toward}, DRealSR~\cite{wei2020component}, and the RAIM challenge~\cite{liang2024ntire}, to construct our Instruct100Set. Specifically, we select and crop images with clear semantic foregrounds to ensure meaningful evaluation. The foreground masks are generated using the pipeline described in Section~\ref{sec:dataset}, ensuring accurate and consistent ROI extraction. The user instructions used in the experiment are in the format of \textit{``make $\{$ target region caption $\}$ clear with $\{$ fidelity level 1 $\}$ and keep other parts clear with $\{$ fidelity level 2 $\}$}."

\noindent \textbf{Evaluation metrics.}
To comprehensively assess the performance of our method, we adopt both reference-based and no-reference metrics, evaluating both target regions and the entire image. The reference-based metrics include PSNR, SSIM~\cite{ssim} (on the Y channel in YCbCr space) and LPIPS \cite{lpips}. The no-reference metrics include MANIQA~\cite{maniqa}, MUSIQ~\cite{ke2021musiq} and CLIPIQA~\cite{clipiqa}. For region-specific evaluation, we compute PSNR and SSIM exclusively within human-specified regions by using the provided GT mask. For other metrics, we zero out pixels outside the target region based on GT mask for computation. This ensures the evaluation focusing on the target regions while being compatible with standard implementations of these metrics.

\begin{table*}
    \centering
    \caption{
        Quantitative evaluation on the instruction following capability of InstructRestore. Experiments are conducted on the Instruct100Set with instruction of "Make \{\ Region caption of target area\}\ clear with \{\ Fidelity Scale \}\ and keep other parts clear with 1."
    }
    \vspace{-0.9em}
    \renewcommand{\arrayrulewidth}{0.2pt} 
    \resizebox{\linewidth}{!}{
        {\fontsize{4}{4.5}\selectfont 
        \begin{tabular}{c|cccccc|cc}
            \hline
            \multirow{2}{*}{Fidelity Scale} & \multicolumn{6}{c|}{Target Area} & \multicolumn{2}{c}{Remaining Area} \\
            \cline{2-9}
            & PSNR\scalebox{0.6}{$\uparrow$} & SSIM\scalebox{0.6}{$\uparrow$} & LPIPS\scalebox{0.6}{$\downarrow$} & CLIPIQA\scalebox{0.6}{$\uparrow$} & MUSIQ\scalebox{0.6}{$\uparrow$} & MANIQA\scalebox{0.6}{$\uparrow$} & PSNR\scalebox{0.6}{$\uparrow$} & SSIM\scalebox{0.6}{$\uparrow$} \\
            \hline
            0.5 & 29.71 & 0.7522 & 0.1610 & 0.6801 & 67.86 & 0.6108 & 31.27 & 0.8949 \\  
            0.7 & 30.37 & 0.8188 & 0.1439 & 0.6931 & 68.23 & 0.6161 & 31.55 & 0.9047 \\  
            0.9 & 30.64 & 0.8494 & 0.1331 & 0.6832 & 67.91 & 0.6091 & 31.61 & 0.9087 \\
            1.1 & 30.73 & 0.8649 & 0.1253 & 0.6659 & 66.92 & 0.5934 & 31.56 & 0.9108 \\
            \hline
        \end{tabular}
        } 
    }
    \vspace{-2mm}
    \label{tab:tuning}
\end{table*}

\begin{figure}[t]
  \centering
  \includegraphics[width=1.0\linewidth]{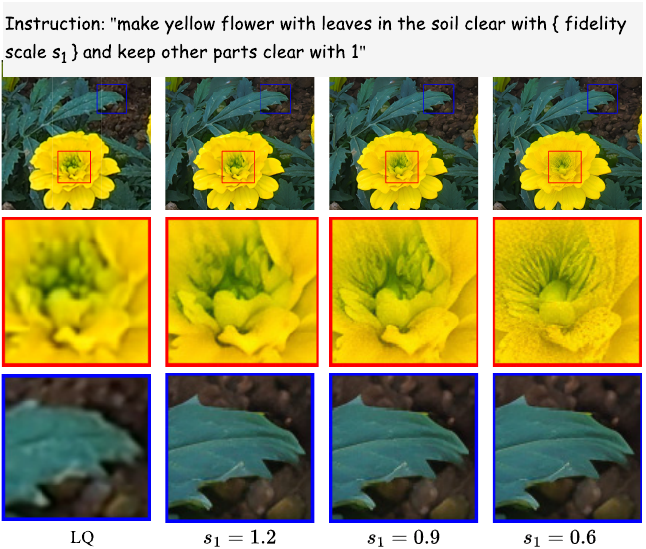}
\vspace{-6mm}
  \caption{Localized enhancement following instruction on real-world test data. The details in flowers are enhanced gradually while the other regions keeping almost unchanged.}
  \label{fig:Demonstration of localized enhancement}
  \vspace{-5mm}
\end{figure}

\noindent \textbf{Localized enhancement with user-instruction}.
We first showcase our method’s ability to perform localized enhancement with user-instructions. By specifying the target region and enhancement strength, our method allows users to explicitly control the balance between data fidelity and generative details. As illustrated in Fig.~\ref{fig:Demonstration of localized enhancement}, by applying different fidelity scale instructions to the flower region, we successfully adjust the level of details in the flower region while keeping other regions (\eg, leaves and soil) largely unchanged. %
To our best knowledge, our method is the first one to allow user-instructed local enhancement.

To quantitatively validate the instruction-following capability of our method, we conduct experiments by varying the enhancement fidelity scales exclusively within the target region while keeping the fidelity scale of the surrounding areas unchanged.
For the surrounding regions, we calculate reference-based metrics to assess their stability.
The quantitative results are shown in  Table~\ref{tab:tuning}. We see that when the fidelity scale is small, the non-reference metrics for the target region are significantly higher, indicating that the method tends to generate more details. As the fidelity scale increases, the reference-based metrics (\eg, PSNR and SSIM) improve, while the non-reference metrics gradually decrease. This demonstrates that the method effectively follows the instructions, transitioning from detail-oriented generation to a more input-faithful reconstruction. Furthermore, the PSNR of the target region varies by $1.02$ db, while the surrounding regions vary only by $0.29$ db. This stark contrast confirms that the enhancement process is localized to the target region, leaving the surrounding areas largely unaffected.


\setlength{\tabcolsep}{3pt}
\begin{table*}\tiny
           \centering

 \caption{
        Quantitative comparison between our InstructRestore method and other methods on Instruct100Set.}
  \vspace{-1.5em}
   \resizebox{\linewidth}{!}{
\begin{tabular}{c|ccccc|cccccc}
\hline
  \multirow{2}{*}{Method} & \multicolumn{5}{c|}{Target Area} & \multicolumn{6}{c}{Full Image} \\
   \cline{2-12}
   & PSNR↑ & SSIM↑ & CLIPIQA↑ & MUSIQ↑ & MANIQA↑ & PSNR↑ & SSIM↑ & LPIPS↓ & CLIPIQA↑ & MUSIQ↑ & MANIQA↑ \\

\hline
  RealESRGAN & \textcolor{blue}{31.69} & \textcolor{blue}{0.9065} & \textcolor{red}{0.7124} & 58.63 & 0.4991 & \textcolor{blue}{27.69} & \textcolor{blue}{0.7871} & 0.3185 & 0.7280 & 60.39 & 0.5030 \\  
  StableSR & 30.36 & 0.8522 & 0.6707 & 65.75 & 0.5915 & 25.39 & 0.7072 & 0.3001 & 0.7072 & 69.19 & \textcolor{blue}{0.6691} \\  
  DiffBIR & 30.95 & 0.8804 & 0.6820 & 66.80 & 0.5971 & 26.64 & 0.6897 & 0.3434 & \textcolor{red}{0.7456} & 69.96 & 0.6609 \\   
  PASD & \textcolor{red}{31.80} & \textcolor{red}{0.9176}  & 0.5724 & 61.02 & 0.5323 & \textcolor{red}{28.37} & \textcolor{red}{0.7893} & \textcolor{red}{0.2590} & 0.5768 & 62.92 & 0.5866 \\
  SeeSR & 30.90 & 0.8788  & 0.6758 & \textcolor{blue}{67.73} & \textcolor{blue}{0.5974} & 26.75 & 0.7324 & 0.2879 & 0.7246 & 71.49 & \textcolor{blue}{0.6691} \\
  SUPIR & 30.74 & 0.8682  & 0.6868 & 62.98 & 0.5655 & 26.29 & 0.6997 & 0.3235 & 0.6840 & 64.40 & 0.6085 \\
  OSEDiff & 30.21 & 0.8657  & 0.6417 & 66.75 & 0.5851 & 26.07 & 0.7340 & \textcolor{blue}{0.2870} & \textcolor{blue}{0.7342} & \textcolor{blue}{71.88} & 0.6635 \\
  Ours & 30.55 & 0.8368  & \textcolor{blue}{0.6887} & \textcolor{red}{68.17} & \textcolor{red}{0.6137} & 25.65 & 0.6999 & 0.3245 & 0.7278 & \textcolor{red}{71.95} & \textcolor{red}{0.6809} \\

\hline
\end{tabular}
}
 \vspace{-1.8em}
\label{tab:Quantitative comparison}
\end{table*}

\begin{figure*}[t]
  \centering
  \includegraphics[width=0.9\linewidth]{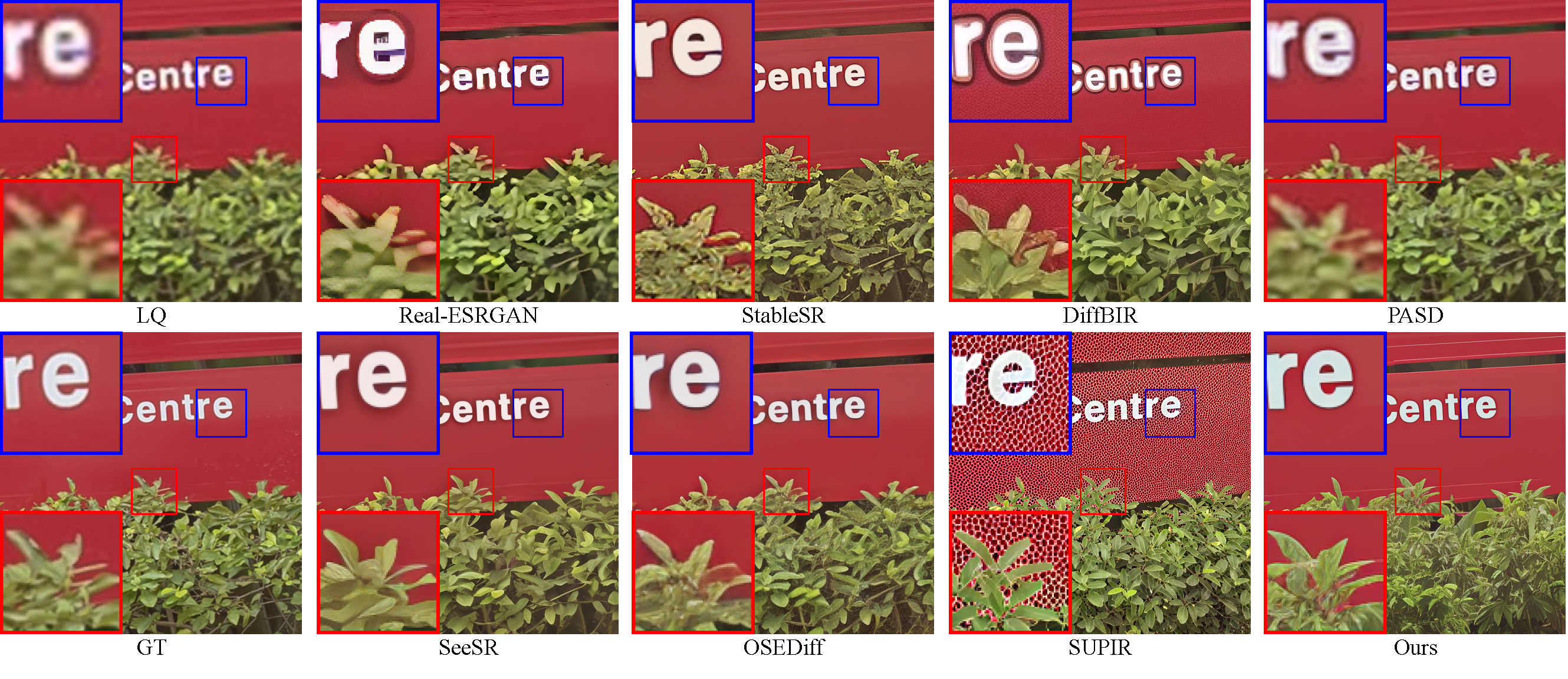}
\vspace{-4mm}
  \caption{Visual comparison of different methods. We set the instruction as \textit{``Make the bush in front of sign clear with 0.5 and keep other parts clear with 0.9"} to keep the fidelity of sign and prioritize detail enhancement of bushes.}
  \label{fig:visual compare in localized enhancement}
  \vspace{-4mm}
\end{figure*}

\noindent \textbf{Comparison with other methods}. 
We then compare InstructRestore with the competing methods. 
For images with heavier degradations, we prioritize stronger generative prior to synthesize more details; for regions with high-frequency and irregular textures (\eg, flowers, brushes), we favor generative enhancement to achieve realistic appearances; while for regions with regular structures (\eg, sign and buildings), a conservative enhancement level is selected to avoid unnatural artifacts. 
As shown in Figure~\ref{fig:visual compare in localized enhancement}, our method can handle well distinct regions, namely the sign and the bushes,  within the same scene. In comparison, methods such as DiffBIR and SUPIR tend to over-enhance the sign, introducing unnecessary artifacts and distortions, while other methods fail to adequately reconstruct the bush, resulting in a smeared and over-smoothed appearance. 

Our method allows adjusting the fidelity scale to meet the specific requirements of each region. For example, for the sign, which requires high fidelity, we set the fidelity scale to $0.9$ for faithful restoration. For the bush, we prioritize detail enhancement with a fidelity scale of $0.5$ to generate richer textures. 
To provide an example of quantitative evaluation, we simply set the fidelity scale for the foreground at $0.8$, while that for other regions to 1. The evaluation results are reported in Table~\ref{tab:Quantitative comparison}. Since this setting prioritizes generative enhancement in target regions to achieve richer details, it shows better no-reference metrics but relatively lower scores in reference metrics that favor strict fidelity preservation. It is important to note that our InstructRestore enables users to adaptively adjust restoration results based on their preferences. The metrics here only serve as an example to demonstrate that our approach can produce visually pleasing results following user instructions. 

\subsection{Results on Images with Bokeh Effects}
In this section, we perform experiments to demonstrate that our method can perform image restoration while preserving bokeh effects and controlling the bokeh intensity.

\noindent \textbf{Test dataset}. 
We construct a test dataset by selecting images from two sources: the EBB! dataset~\cite{ignatov2020rendering} and images with bokeh background carefully curated from Pixabay~\cite{pixabay}. We select $70$ images from them with distinct semantic foregrounds and bokeh background. Masks are generated for the foreground regions to precisely define the ROI, based on which the images are center-cropped to ensure a consistent resolution of $512$ $\times$ $512$. Subsequently, we apply Real-ESRGAN~\cite{wang2021real} degradations to generate LQ and GT image pairs for evaluation. To support instructed interaction, we leverage the masks and GT images to generate foreground descriptions using the pipeline illustrated in Section~\ref{sec:dataset}. The user instructions are formatted as \textit{``Make $\{$foreground description$\}$ clear with $\{$enhancement level$\}$ and keep the bokeh blur of other parts with the $\{$bokeh level$\}$." } 

\noindent \textbf{Evaluation Metrics.} We compute reference-based metrics for full image and background regions. In addition, we employ D-DFFNet~\cite{jin2023depth}, a model for detecting blurred background, to generate the background mask and compute the Intersection-over-Union (IoU) with GT of background mask as a measure of bokeh preservation performance.

\begin{figure}[htbp]
    \centering
    \includegraphics[width=\linewidth]{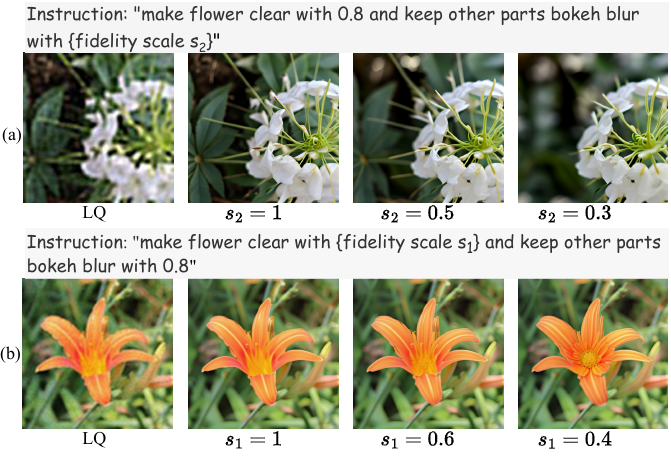}
    \vspace{-7mm}
    \caption{Control of bokeh effect and foreground enhancement. (a) Restoration with controlled bokeh effect while restoring foreground. (b) Restoration with varying foreground enhancement levels while preserving background bokeh.} 
    \label{fig:Demonstration of controlled bokeh and foreground} 
    \vspace{-3mm}
\end{figure}

\noindent \textbf{Control of bokeh effect}. Our method allows users to specify the desired intensity of bokeh effects and foreground enhancement level via instructions. As illustrated in Fig. \ref{fig:Demonstration of controlled bokeh and foreground} (a), our method successfully adjusts the background blur based on user instructions, simulating varying depth-of-field effects while maintaining the sharpness and details in the foreground. More importantly, the adjusted blur is not merely a uniform increase in blur intensity. It faithfully replicates the circular light spots of realistic bokeh, mimicking the optical effects produced by high-quality digital single-lens reflex (DSLR) cameras. %
In addition to specifying the intensity of bokeh effects, users can further specify the enhancement strength for the foreground, achieving flexible control on the level of details in focal regions. As illustrated in Fig. \ref{fig:Demonstration of controlled bokeh and foreground} (b), the fine details in the flower stamens become more pronounced as the instruction changes. Note that such a feature is not supported by existing restoration methods.

\begin{figure*}[htbp]
   \centering
   \includegraphics[width=\linewidth]{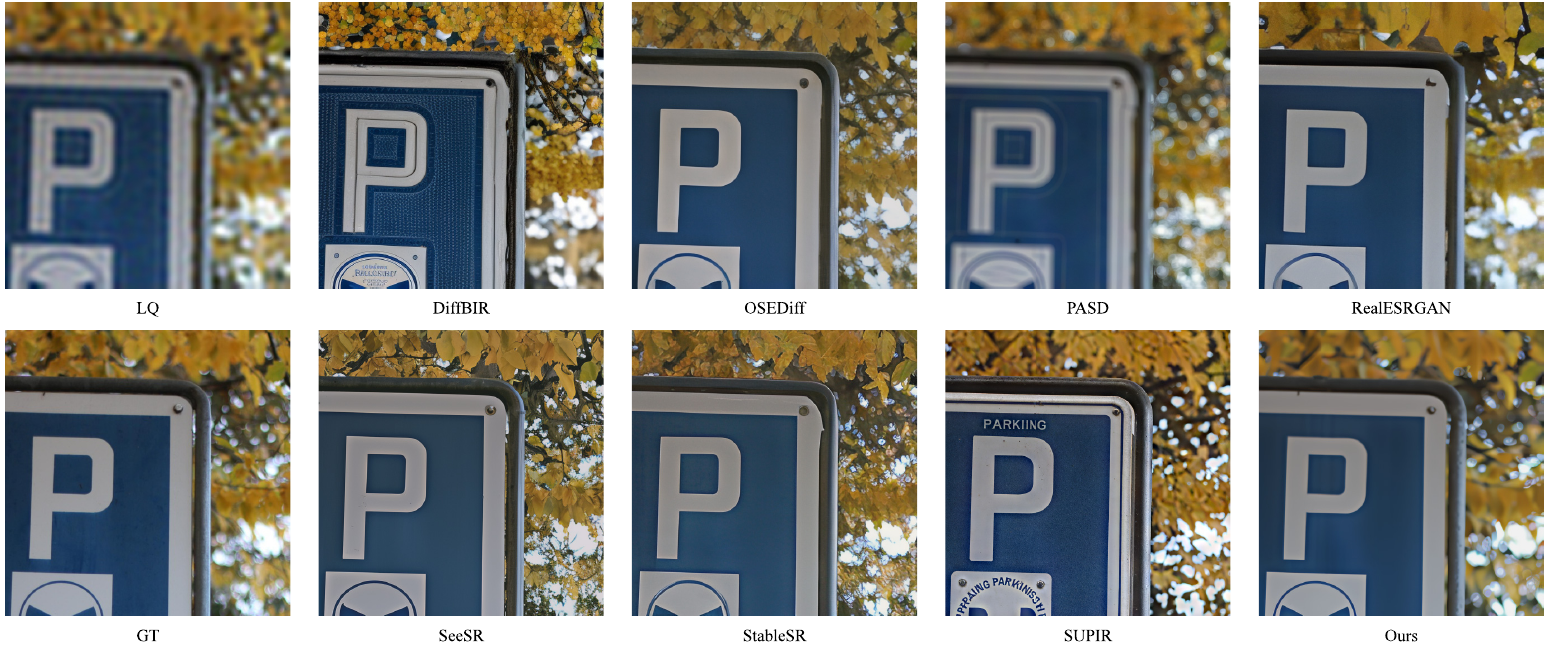}
   \vspace{-4mm}
    \caption{Visual Comparison of bokeh preservation results between the compared methods.} 
    \label{fig:Visual Comparison of bokeh preservation with baselines} 
    \vspace{-4mm}
\end{figure*}

\setlength{\tabcolsep}{3pt}
\begin{table}\tiny
           \centering

 \caption{
        Quantitative comparison on Bokeh testset}
  \vspace{-3mm}
   \resizebox{\linewidth}{!}{
\begin{tabular}{c|ccc|ccc}
\hline
  \multirow{2}{*}{Method} & \multicolumn{3}{c|}{Background} & \multicolumn{3}{c}{Full Image} \\
   \cline{2-7}
   & PSNR↑ & SSIM↑ & Bokeh IoU↑ & PSNR↑ & SSIM↑ & LPIPS↓ \\

\hline
  RealESRGAN & 30.86 & 0.8305 & 0.7203 & 23.69 & 0.7060 & 0.3700 \\  
  StableSR & 30.24 & 0.8049 & 0.7405 & 22.55 & 0.6305 & 0.3965   \\  
  DiffBIR & 30.46 & 0.8017 & 0.6289 & 22.20 & 0.5943 & 0.4415  \\   
  PASD & \textcolor{red}{31.87} & \textcolor{blue}{0.8453}  & \textcolor{blue}{0.8234} & \textcolor{blue}{24.27} & \textcolor{blue}{0.7280} & \textcolor{blue}{0.3523} \\
  SeeSR & 30.42 & 0.8149  & 0.7580 & 22.95 & 0.6652 & 0.3677 \\
  SUPIR & 29.92 & 0.7847  & 0.7739 & 21.21 & 0.5745 & 0.4375 \\
  OSEDiff & 29.89 & 0.8175  & 0.7990 & 22.58 & 0.6707 & 0.3609 \\
  Ours & \textcolor{blue}{31.46} & \textcolor{red}{0.8462} & \textcolor{red}{0.8482} & \textcolor{red}{24.69}  & \textcolor{red}{0.7437}  & \textcolor{red}{0.3394}  \\

\hline
\end{tabular}
}
 \vspace{-5mm}
\label{tab:Quantitative comparison on bokeh testset}
\end{table}

\noindent \textbf{Comparison with other methods}. Since the foreground semantics in EBB! mainly include objects like cars and road signs requiring high fidelity, we set the foreground enhancement strength to $1.0$. For simplicity and fairness, the bokeh fidelity scale is also set to a default value of $1$, representing the weakest depth-of-field effect. The quantitative comparison results are presented in Tab.~\ref{tab:Quantitative comparison on bokeh testset}. Our method demonstrates significantly better performance in fidelity-oriented metrics compared to competing methods, reflecting its ability to accurately approximate the GT’s bokeh characteristics. In contrast, existing methods fail to preserve bokeh effects, leading to deviations from the GT. Visual comparisons are provided in the Fig.~\ref{fig:Visual Comparison of bokeh preservation with baselines}. Except for PASD~\cite{yang2024pixel} and SUPIR~\cite{yu2024scaling}, other methods restore sharp details in the background. However, PASD fails to recover clear foreground details, while SUPIR over-enhances the image, generating numerous artifacts that are easily discernible to the naked eye. In contrast, our method maintains foreground fidelity while preserving the background bokeh effect.
\section{Conclusion}
We presented InstructRestore, the first framework for region-customized image restoration guided by human instructions. To support this task, we designed a scalable data annotation engine and constructed a dedicated dataset comprising 536,945 triplets, each containing a high-quality image, the region mask and region caption. Building on this dataset, we developed an InstructRestore model that parsed human instructions to achieve region-specific restoration. Our framework allowed users to apply distinct enhancement intensities to different regions and adjust background bokeh effects. By enabling fine-grained control via user instructions, our work advanced research in interactive image restoration and enhancement techniques.

\noindent \textbf{Limitations.} While InstructRestore offers a baseline for region-customized restoration guided by human instructions, it has several limitations. Currently, it lacks support for instance-level object specification, which requires instance-level masks and captions. Moreover, users must adhere to a predefined instruction template. Furthermore, although our method achieves competitive results, it focuses more on localized customization, while it deserves further exploration of global quality optimization. Addressing these limitations would greatly enhance the applicability and performance of user-instructed image restoration.

{
    \small
    \bibliographystyle{ieeenat_fullname}
    \bibliography{main_v2}
}
\appendix
\onecolumn
\section{Sumpplementaty File}

In the supplementary file, we provide the following materials:

\begin{itemize}
    \item \textbf{Details of Prompt Tuning with Qwen}:  
    We provide the implementation details of using Qwen for prompt tuning as discussed in Section~3 of the main paper, including the model setup, prompt design.
    \item \textbf{More Visual Results for Instruction Following}:  
    We showcase additional visual results demonstrating the localized restoration on images from the Instruct100set based on human instructions (referring to Section 5 in the main paper).
    \item \textbf{More Visual Comparisons on Instruct100Set}:  
    We present more visual comparisons with other methods on the Instruct100Set (referring to Section 5 in the main paper).
    \item \textbf{More Visual Comparisons on Bokeh Dataset}:  
    We present more visual comparisons on the Bokeh Dataset, illustrating the model’s performance for the bokeh effect preservation (referring to Section 5 in the main paper).
\end{itemize}

\subsection{Details of Prompt Tuning with Qwen}
\label{sec:Details of Prompt Tuning with Qwen}
\begin{figure*}[h]
  \centering
  \includegraphics[width=0.9\linewidth]{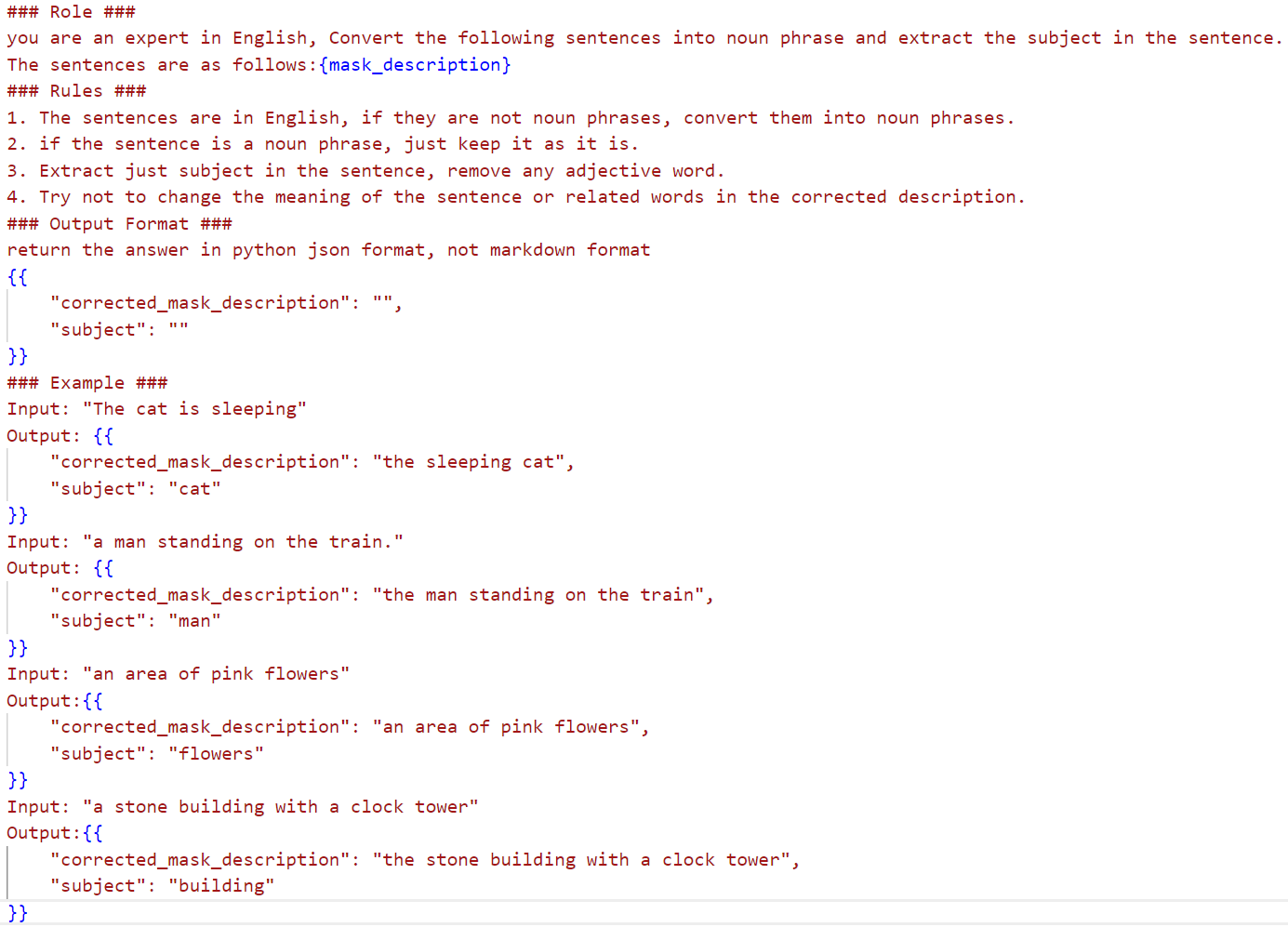}
  \vspace{-3mm}
  \caption{The designed prompts for Qwen to enable parsing and reformatting the initial region descriptions.}
  \label{fig:prompt}
\end{figure*}

In Section~3 of the main paper, we mentioned that the initial masks and region descriptions obtained from Semantic-SAM~\cite{li2024segment} and Osprey~\cite{yuan2024osprey} cannot meet the requirements for our training data. To address this, we employed a Large Language Model (LLM), specifically Qwen~\cite{qwen}, to reformat the region captions into a noun phrase structure and extract the subject from the descriptions, which facilitates the merging of masks and region captions with identical semantics. As illustrated in Figure~\ref{fig:prompt}, we designed a specific prompt for this purpose. In the first round of parsing, we utilized the Qwen-7B~\cite{qwen} model to ensure time efficiency.

However, due to the inherent randomness of LLM outputs, issues such as spelling errors or repetitive text (\ie, ``parroting") can arise. We found that the frequency of the parsed subjects could serve as a useful indicator of their correctness. Given that the number of subjects parsed in real-world scenarios typically does not exceed 200, subjects with frequencies ranked beyond the top 200 are likely to be parsing errors. For these instances, we revisited the original data and implemented the parsing process using the larger Qwen-72B~\cite{qwen} model. This iterative procedure was repeated twice, resulting in a total of three iterations to refine and finalize our annotations.

\subsection{More Examples of Localized Enhancement with User-Instruction}
\begin{figure*}[t]
  \centering
  \includegraphics[width=0.6\linewidth]{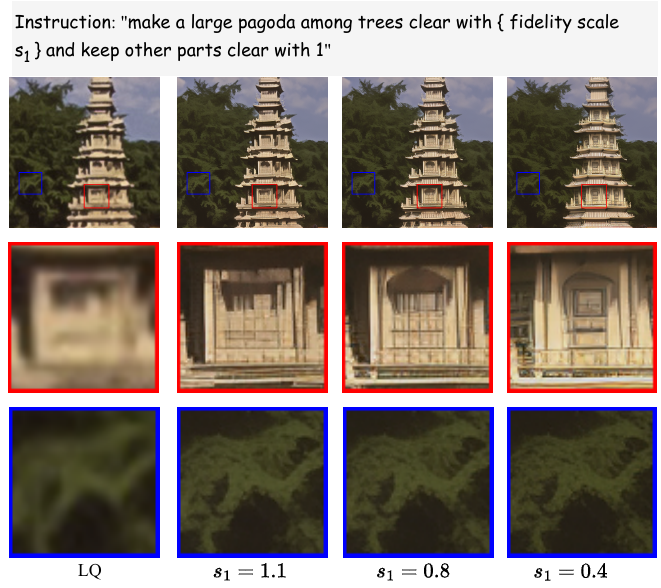}
  \vspace{-3mm}
  \caption{The varied localized enhancement for pagoda.}
  \label{fig:real_tuning_supp1}
\end{figure*}

\begin{figure*}[t]
  \centering
  \includegraphics[width=0.6\linewidth]{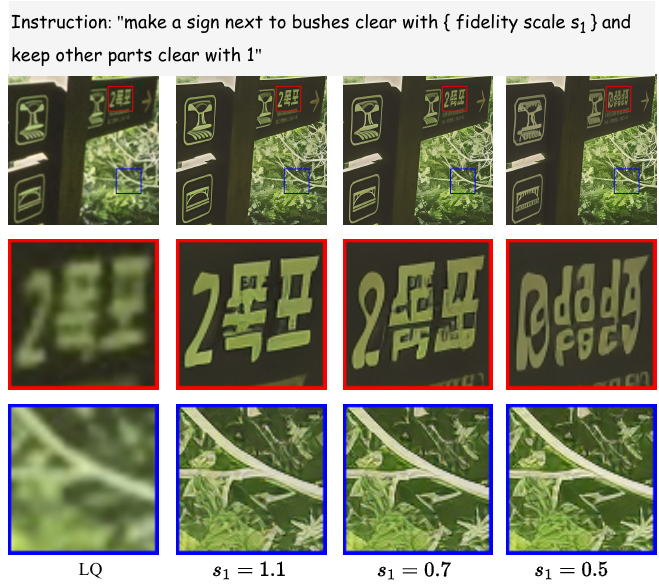}
  \vspace{-3mm}
  \caption{The varied localized enhancement for sign.}
  \label{fig:real_tuning_supp2}
\end{figure*}

\begin{figure*}[t]
  \centering
  \includegraphics[width=0.6\linewidth]{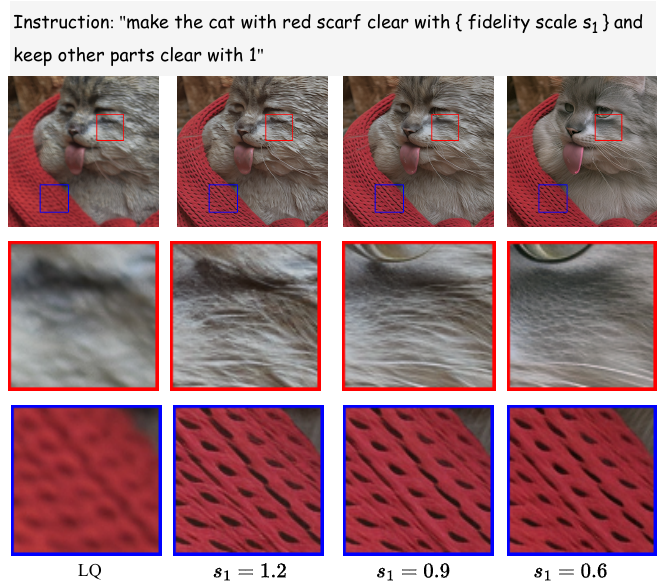}
  \vspace{-3mm}
  \caption{The varied localized enhancement for cat.}
  \label{fig:real_tuning_supp3}
\end{figure*}

In Figure 4 of Section~5.2 of the main paper, we used flowers as an example to demonstrate the effectiveness of our method in achieving region-specific restoration based on user instructions. %
In this section, we provide more examples of various semantic scenes to further validate the comprehensive semantic coverage of our dataset and the effectiveness of our method in following user instructions. In Fig.~\ref{fig:real_tuning_supp1}, as the fidelity scale decreases, the texture of the pagoda becomes increasingly detailed, with clearly recognizable windows emerging. Meanwhile, the restoration results of the surrounding trees remain unchanged. In Fig.~\ref{fig:real_tuning_supp2}, the text on the sign closely resembles the input when the fidelity scale is set to 1.1. As the fidelity scale decreases, the restored patterns begin to deviate from the input. In contrast, the restoration results of the surrounding plants remain unchanged. In Fig.~\ref{fig:real_tuning_supp3}, when the fidelity scale is set to 1.2, the cat's fur appears clustered. As the fidelity scale decreases, the fur becomes more refined. However, the facial features of the cat, especially the eyes, increasingly deviate from the input, demonstrating that the fidelity indeed decreases as instructed. %
These examples highlight the capability of our dataset to support the processing of common semantic categories, as well as the effectiveness of our method in understanding instructions, localizing target regions, and adaptively adjusting the restoration results.

\subsection{More Visual Comparisons on Instruct100Set}
\begin{figure*}[t]
  \centering
  \includegraphics[width=0.9\linewidth]{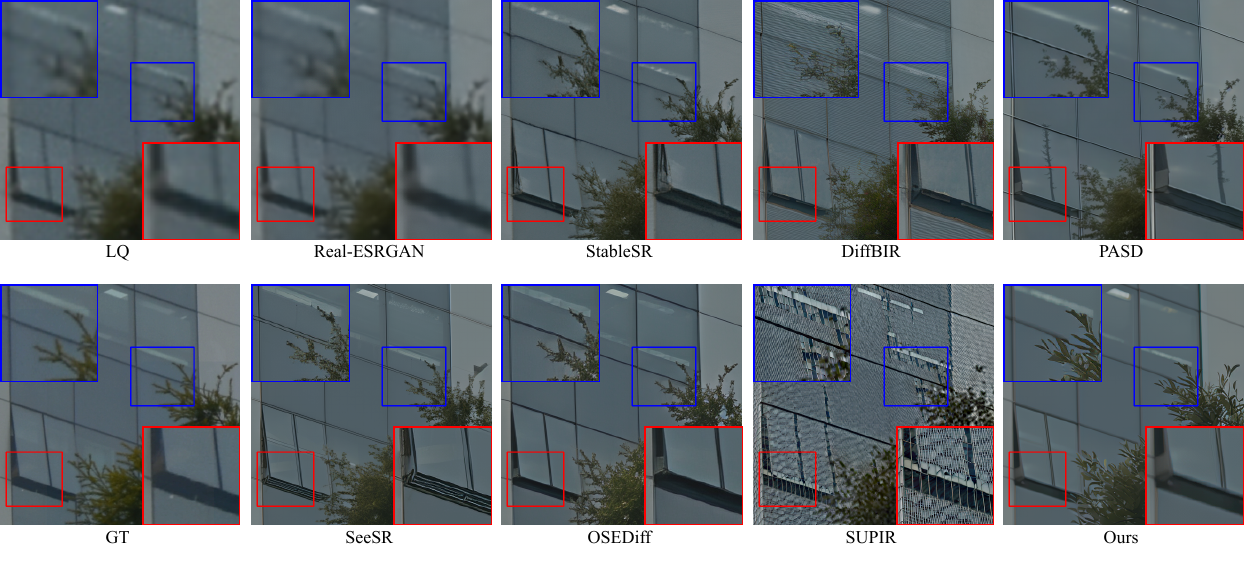}
  \vspace{-3mm}
  \caption{Example of visual comparison on Instruct100Set.}
  \label{fig:localized_compare1}
\end{figure*}

\begin{figure*}[t]
  \centering
  \includegraphics[width=0.9\linewidth]{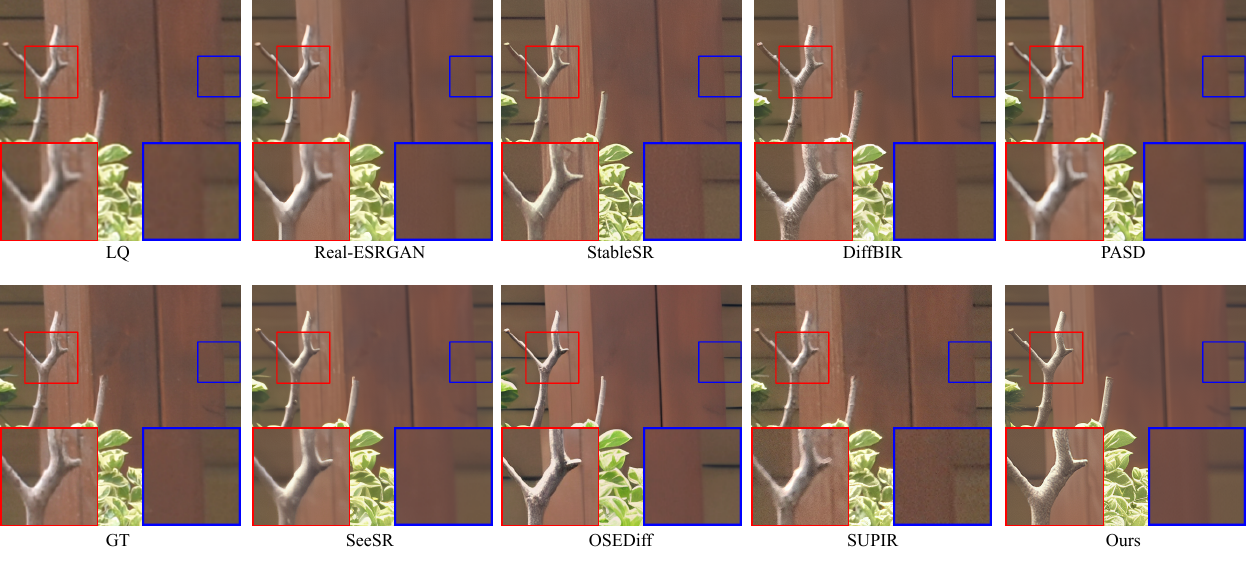}
  \vspace{-3mm}
  \caption{Example of visual comparison on Instruct100Set.}
  \label{fig:localized_compare2}
\end{figure*}

\begin{figure*}[t]
  \centering
  \includegraphics[width=0.9\linewidth]{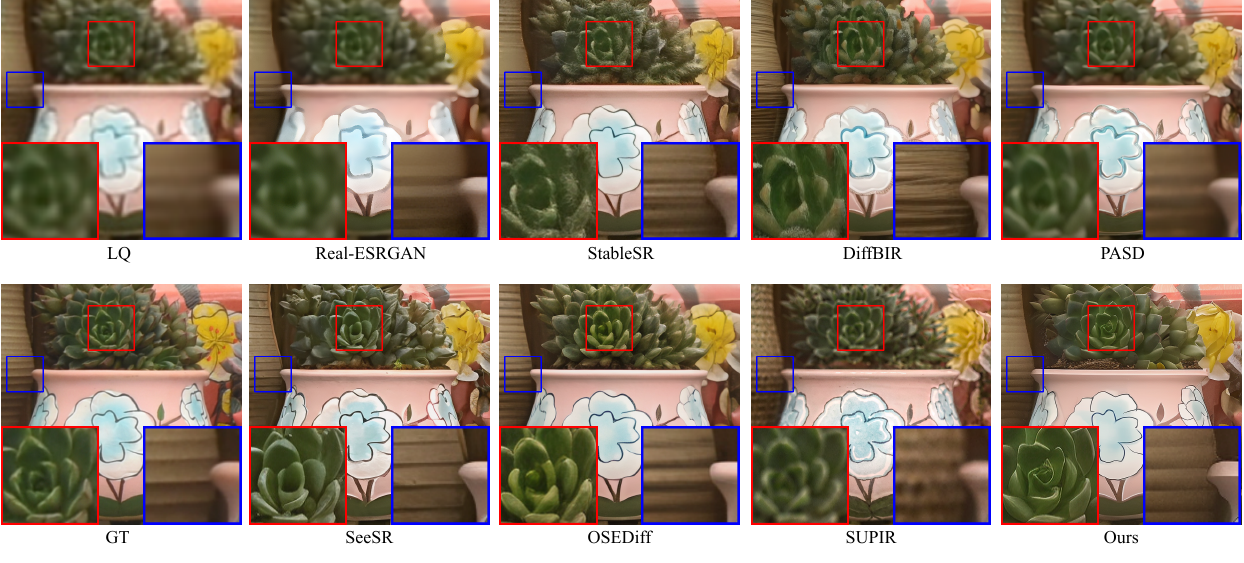}
  \vspace{-3mm}
  \caption{Example of visual comparison on Instruct100Set.}
  \label{fig:localized_compare3}
\end{figure*}

In this section, we provide additional examples comparing our method with existing restoration approaches on the Instruct100Set. As shown in Fig.~\ref{fig:localized_compare1}, Real-ESRGAN~\cite{wang2021real} fails to restore fine details. Except for OSEDiff~\cite{wu2025one}, other methods generate unnecessary artifacts in the windows of the building. However, OSEDiff restores the indoor lights as prominent white marks. In addition, the foliage restored by these methods exhibits a smeared appearance. In contrast, our method ensures the fidelity of the building while enhancing the details of plants, resulting in a more realistic overall appearance. In Fig.~\ref{fig:localized_compare2}, our method restores more intricate details in the tree branches while maintaining background fidelity. In contrast, other methods, except for DiffBIR~\cite{lin2024diffbir}, lack realistic details in the restoration of branches. However, similar to StableSR~\cite{wang2024exploiting} and SUPIR~\cite{yu2024scaling}, DiffBIR exhibits incomplete denoising in the background regions. In Fig.~\ref{fig:localized_compare3}, none of the other methods successfully restores the fleshy leaves of the succulent plant, which can be seen in the input. Additionally, in the results of StableSR and DiffBIR, the texture details of the adjacent clay pot are overly cluttered.

\subsection{More Visual Comparisons on Bokeh Testset}
\begin{figure*}[t]
  \centering
  \includegraphics[width=0.9\linewidth]{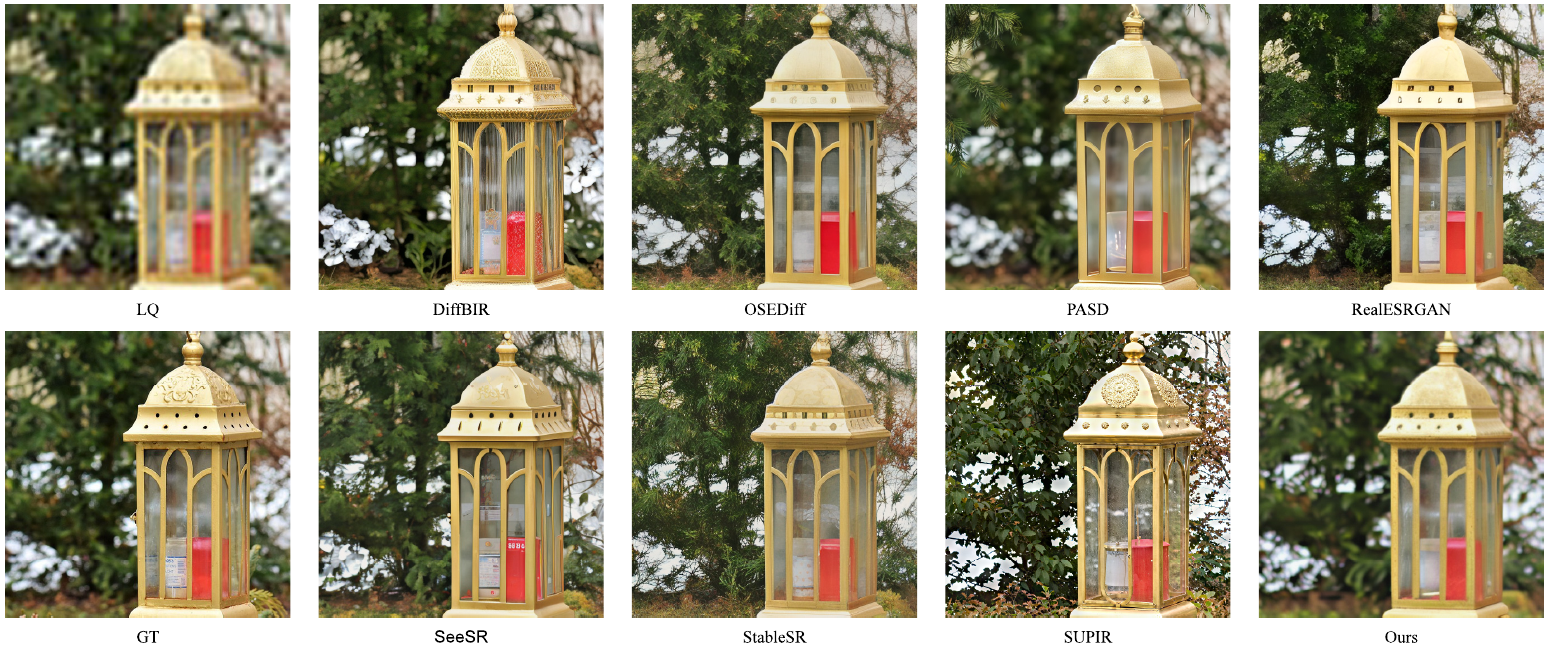}
  \vspace{-3mm}
  \caption{Example of visual comparison on Bokeh Testset.}
  \label{fig:2nd_bokehcompare}
\end{figure*}

\begin{figure*}[t]
  \centering
  \includegraphics[width=0.9\linewidth]{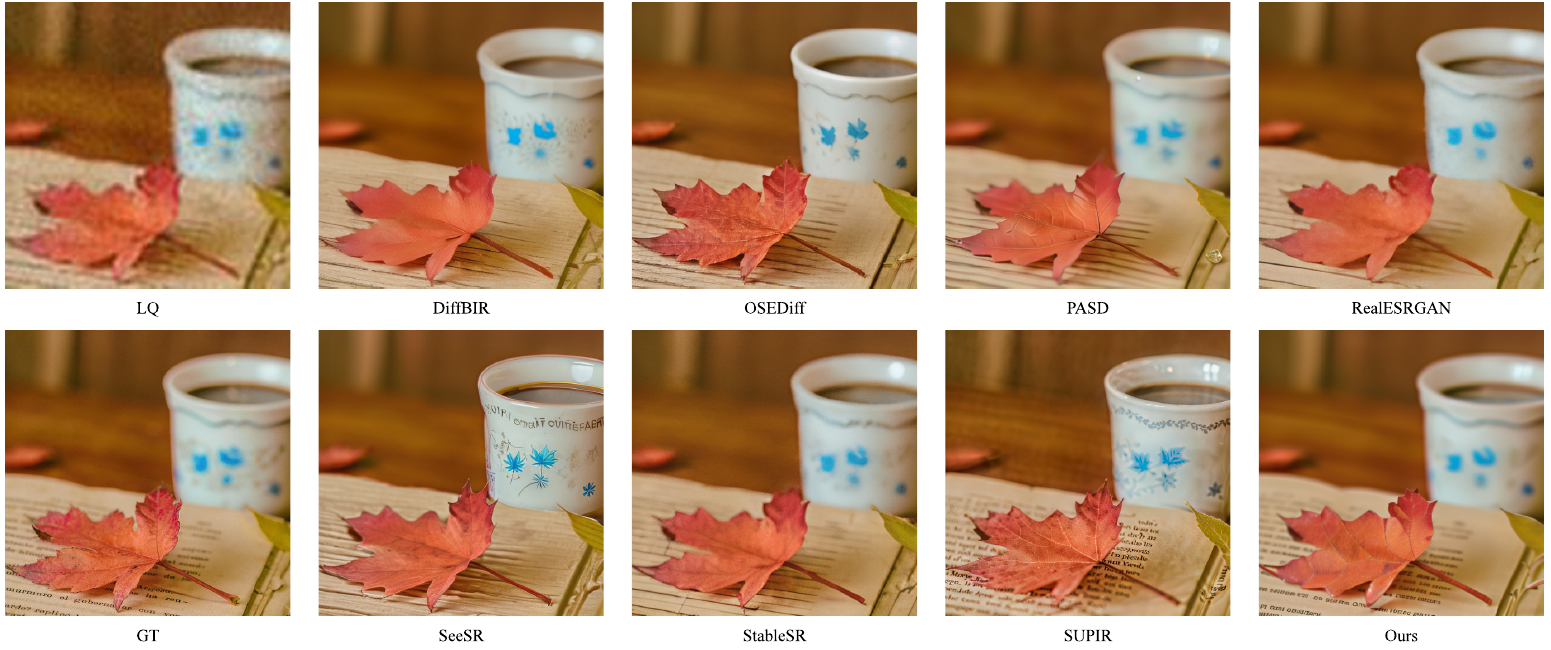}
  \vspace{-3mm}
  \caption{Example of visual comparison on Bokeh Testset.}
  \label{fig:3rd_bokehcompare}
\end{figure*}

\begin{figure*}[t]
  \centering
  \includegraphics[width=0.9\linewidth]{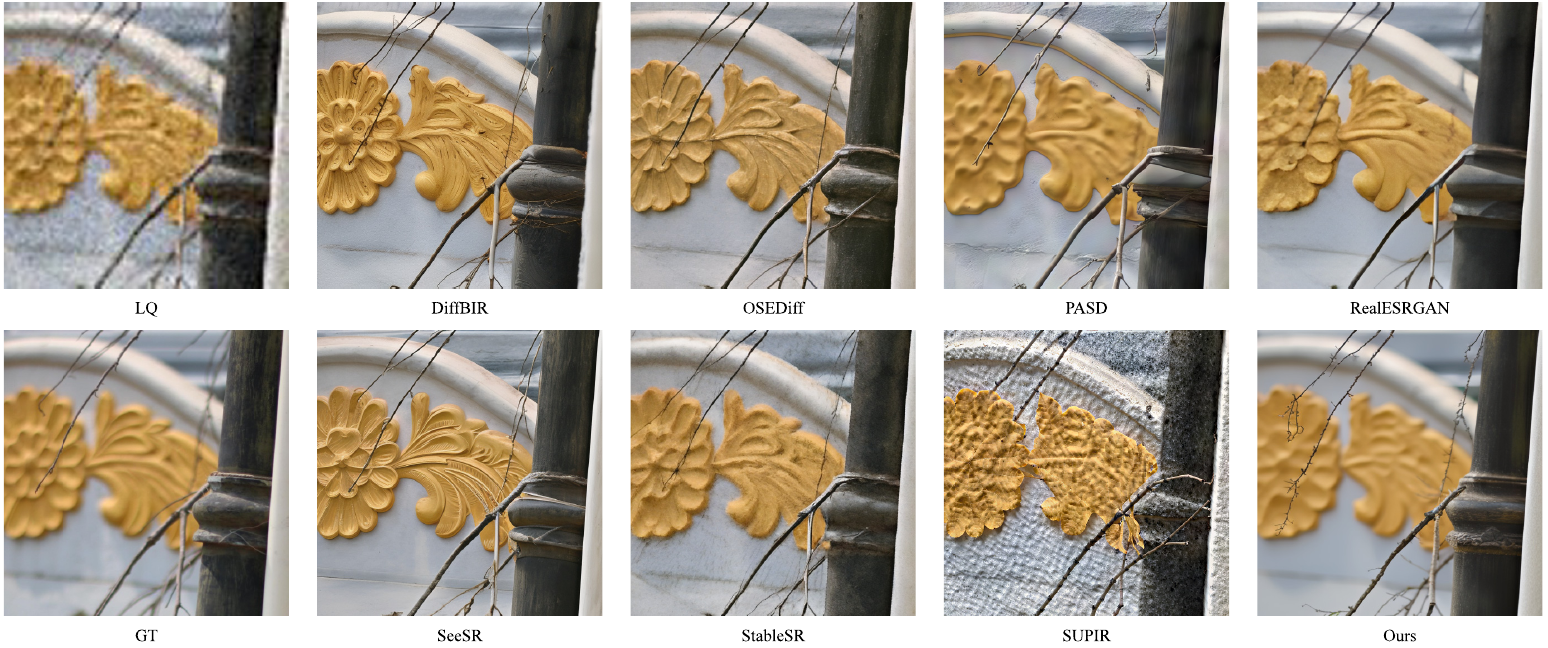}
  \vspace{-3mm}
  \caption{Example of visual comparison on Bokeh Testset.}
  \label{fig:4th_bokehcompare}
\end{figure*}

In Section 5.3 of the main paper, we highlighted that our method outperforms existing approaches in preserving the natural background bokeh effect of scenes. In this section, we provide more visual comparisons to demonstrate this advantage. For a fair comparison, we set the fidelity scale for background bokeh blur to 1 in the instruction, which corresponds to the minimal blurring effect in our method. As shown in Fig.~\ref{fig:2nd_bokehcompare}, DiffBIR restores the blurred light spot in the lower-left corner into a flower, while PASD recovers partial vegetation in the upper-right corner. Other methods restore sharp details to varying degrees. In Fig.~\ref{fig:3rd_bokehcompare}, although StableSR, PASD, and RealESRGAN preserve the background bokeh effect, the texture of the leaves is less clear compared to our results. In addition, the books beneath the leaves are restored to resemble wooden planks, indicating that their foreground fidelity is not as well maintained as ours. In Fig.~\ref{fig:4th_bokehcompare}, only our method keeps the background bokeh blur. Overall, these visual comparisons further demonstrate the effectiveness of our method in preserving the bokeh effect.

\end{document}